\pdfoutput=1

\documentclass[11pt]{article}

\usepackage[final]{acl}

\usepackage{times}
\usepackage{latexsym}
\usepackage{tikz}
\usepackage{hyperref}
\usepackage[edges]{forest}
\usepackage{graphicx}
\usepackage{algorithm}
\usepackage[noend]{algpseudocode}
\algnewcommand{\LeftComment}[1]{\Statex{}$\triangleright$~#1}
\usepackage{multicol}
\usepackage{rotating}
\usetikzlibrary{trees,shapes,calc,arrows.meta,positioning}
\usepackage{makecell}
\usepackage[T1]{fontenc}
\usepackage[utf8]{inputenc}
\usepackage{microtype}
\usepackage{amsmath}
\usepackage{amssymb}
\usepackage{booktabs}
\usepackage{multirow}
\usepackage{colortbl}
\usepackage{adjustbox}
\usepackage{enumitem}
\usepackage{xcolor}
\usepackage{pifont}
\usepackage{amsmath,amssymb,amsthm}
\usepackage{booktabs}
\usepackage{array}
\usepackage{tabularx}
\usepackage{longtable}
\usepackage{multirow}
\usepackage{enumitem}
\usepackage{xcolor}
\usepackage{graphicx}
\usepackage{xspace}
\definecolor{branch1}{HTML}{2E86C1}
\definecolor{branch2}{HTML}{27AE60}
\definecolor{branch3}{HTML}{E67E22}
\definecolor{branch4}{HTML}{8E44AD}
\definecolor{branch5}{HTML}{C0392B}
\definecolor{leafbg}{HTML}{F0F4F8}

\title{The Rise of Verbal Reinforcement Learning}

\author{
  Kshitij Tayal\textsuperscript{1}, 
  Arun Sharma\textsuperscript{2}, 
  Genta Indra Winata\textsuperscript{1}, 
  Anirban Das\textsuperscript{1},
  Sambit Sahu\textsuperscript{1} \\
  \textsuperscript{1}AI Foundations, Capital One \\
  \textsuperscript{2}University of Minnesota, USA \\
  \texttt{\{kshitij.tayal, genta.winata, anirban.das, sambit.sahu\}@capitalone.com} \\
  \texttt{arunshar@umn.edu}
}

\begin{document}
\maketitle
\pagestyle{plain}
\thispagestyle{plain}

\begin{abstract}
Natural language is emerging as a primary feedback channel for improving language agents, capable of conveying intent, preferences, and causal structure in forms interpretable by both humans and modern language models. We call this paradigm Verbal Reinforcement Learning (VRL) and offer the first unified account of it. We organize the field around a single axis, \textit{when} verbal feedback takes effect in an agent's lifecycle and \textit{what} it modifies, yielding three pillars: (1) \textbf{Language as Grounding Signal}, where language defines the task itself by specifying goals, states, and reward structures; (2) \textbf{Language as Deliberative Feedback}, where natural language guides reasoning at test time without the need to update model parameters; (3) \textbf{Language as Learning Signal}, where language-based feedback shapes model parameters through training. Within each pillar, we synthesize representative work, distinguish key subcategories of approaches, and outline the distinct role language plays in shaping agent behavior. Together, this taxonomy shows how verbal reinforcement is reshaping agent development, while also defining the challenges and opportunities for building more capable and aligned agents.
\end{abstract}

\section{Introduction}

\begin{figure}[!th]
     \centering    
    \includegraphics[width=1.1\linewidth]{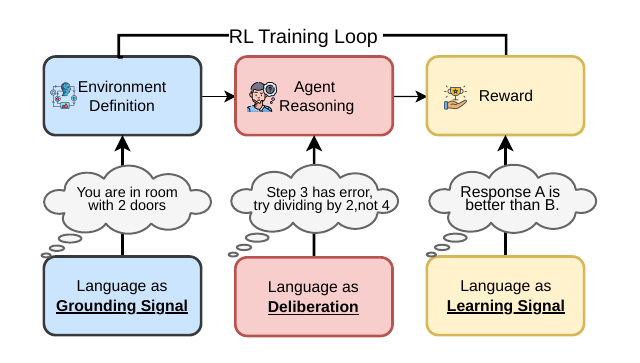}
   \caption{\footnotesize  Language plays three complementary roles in agent's development lifecycle: grounding the environment, scaffolding intermediate reasoning, and providing learning signal for further training.}
    \label{fig:VRLrepresentation}
\end{figure}

\begin{figure*}[!th]
    \centering    
    \includegraphics[width=\textwidth]{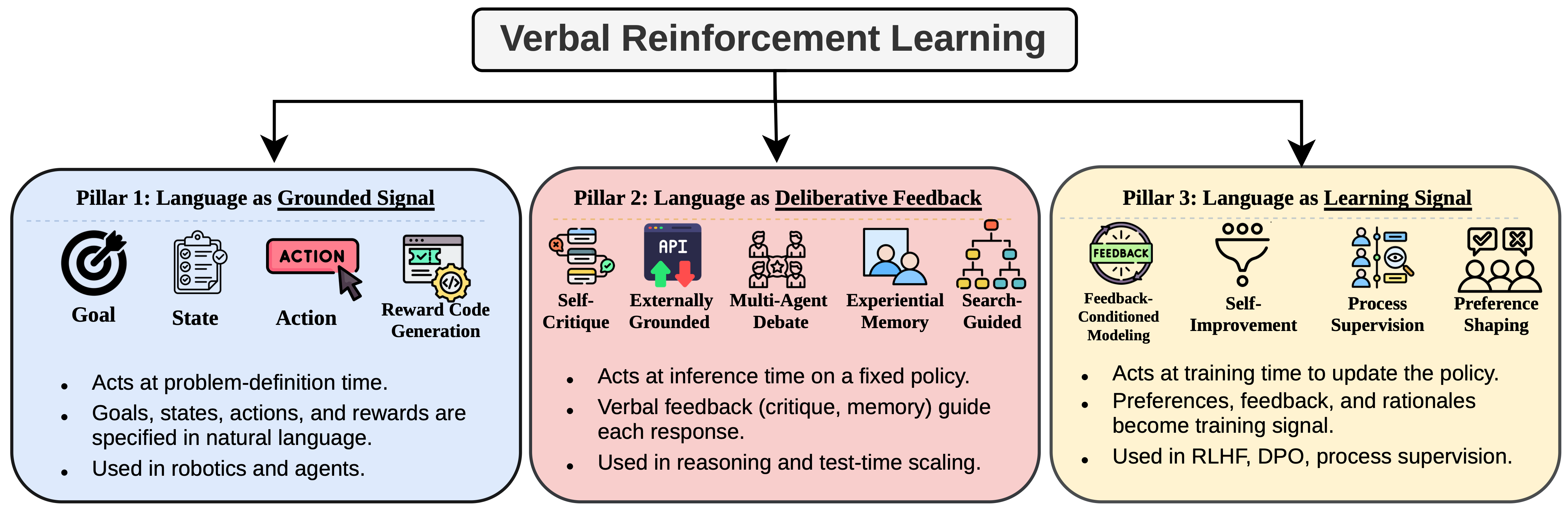}
    \caption{\footnotesize Our proposed taxonomy of Verbal Reinforcement Learning. Language serves three distinct functional roles: \textit{grounding signal}, which structures environment interaction; \textit{deliberative feedback}, which operates at inference time to refine reasoning; and \textit{learning signal}, which drives parameter updates at training time.
    }
    \label{fig:taxonomy}
\end{figure*}

Classical reinforcement learning has achieved remarkable successes by coupling agents with well-specified environments, action spaces, reasoning mechanisms, and task-specific reward functions. This framework has enabled agents to master Atari games \citep{mnih2015human}, defeat world champions at Go \citep{silver2016mastering}, optimize chip placement \citep{mirhoseini2021graph}, and control complex physical systems \citep{degrave2022magnetic}. However, each success rests on heavy domain-specific engineering: the environment must be formalized, the interaction space defined, and the reward function must closely capture the intended behavior \citep{vamplew2021scalarrewardenoughresponse}. Designing such specifications is a long-standing bottleneck \citep{amodei2016concrete}, especially in settings where tasks and goals are ambiguous, context-dependent, and difficult to encode directly.

The emergence of large language models \citep{brown2020language} has introduced an alternative supervisory channel: natural language. Instead of requiring all task intent to be predefined, verbal feedback enables on-the-fly communication of corrections, preferences, and constraints in a manner that agents are increasingly able to interpret and implement. Modern large language models, particularly frontier models \citep{achiam2023gpt}, can accurately describe a wide range of phenomena. These capabilities include describing environments \citep{wang2023voyager}, identifying bugs in code \citep{mcaleese2024criticgpt}, critiquing generated outputs \citep{madaan2023selfrefine}, and translating human preferences into actionable training signals \citep{hong2024orpo, meng2024simpo}. Similar to how self-supervised learning \citep{devlin2019bertpretrainingdeepbidirectional, chen2020simple} expanded the use of unlabeled data in model development, verbal feedback is emerging as an important source of supervision for language-model-based agents.

We refer to this family of methods as \textbf{Verbal Reinforcement Learning (VRL)}: \textit{a paradigm in which an agent receives natural-language feedback---whether self-generated, provided by a human, or produced by an external tool or model---and uses it to improve its behavior and guide future decision-making}. Feedback may be used directly as text or converted into a compressed training signal for parameter updates. The term was introduced by \citet{shinn2023reflexion} to describe self-reflective agents (specific instance of what we call deliberative feedback); here, it is adopted more broadly to encompass any method in which verbal feedback shapes an agent's outputs, learned policy, or problem specification.

Early results demonstrate that verbal feedback can drive substantial improvements across diverse settings. \citet{ahn2022can} used language to ground high-level instructions in physical affordances, enabling real-world robotic planning. \citet{shinn2023reflexion} achieved 91\% pass@1 on HumanEval through verbal self-reflection alone, and \citet{madaan2023selfrefine} showed that iterative verbal critique yields substantial gains across diverse generation tasks. \citet{ouyang2022training} demonstrated that a 1.3B-parameter model trained on verbal preference judgments outperformed the 175B GPT-3 baseline, a 130$\times$ size disadvantage overcome by richer supervision. These are not isolated demonstrations and publications on arXiv referencing ``self-refine,'' ``self-correction,'' ``verbal feedback,'' and ``language-feedback alignment'' have grown from a handful in early 2020 to several hundred today, spanning multiple domains including coding assistants \citep{yang2024swe, jimenez2024swebenchlanguagemodelsresolve}, scientific discovery \citep{m2024augmenting, lu2024ai}, robotics \citep{wang2023voyager}, mathematical reasoning \citep{hatamizadeh2026igrpo, liu2026dual, shi2025ssr}, clinical decision-making \citep{huang2025medreflect, zhou2025enhancing} and education \citep{qian2025dean, ahtisham2026ai}.

These methods differ in their mechanisms (Figure~\ref{fig:VRLrepresentation}): some operate during inference, others modify training data, and some redefine the task itself . However, they share a common characteristic: the use of natural language to guide, refine, and ground agent behavior. Yet no existing framework unifies them. Prior surveys have examined language-conditioned policies in text-based environments \citep{luketina2019survey}, the bidirectional synergies between LLMs and RL \citep{pternea2024rl}, LLM self-correction in isolation \citep{kamoi2024can}. In contrast, our survey focuses on the functional role of \emph{verbal feedback} as a first-class signal for improving agent, organized along a single axis: \emph{when} natural language enters the agent's lifecycle and \emph{what} it modifies. We formalize this axis into a three-pillar taxonomy in Section~\ref{sec:taxonomy}. This paper makes following contributions:

 \begin{itemize}

    \item We present the first comprehensive survey of Verbal Reinforcement Learning, defining the paradigm broadly and proposing a three-pillar taxonomy (Section~\ref{sec:taxonomy}) that organizes the field by \emph{when} language acts in the agent lifecycle.
    \item Within each pillar, we synthesize representative methods, identify cross-cutting challenges and outline future directions for developing robust, capable and aligned Verbal RL agents.

\end{itemize}

We organize the paper as follows. Section~\ref{sec:taxonomy} introduces the three-pillar taxonomy and a working example showing how the pillars operate together in a single coding agent. Sections~\ref{sec:pillar1}--\ref{sec:pillar3} develop each pillar of the taxonomy in detail: language as grounding signal, as deliberative feedback, and as learning signal, respectively. Finally, section~\ref{sec:discussion} presents cross-cutting insights and outlines future research directions, while section~\ref{sec:conclusion} provides the conclusion.

\section{Taxonomy}
\label{sec:taxonomy}
We organize VRL methods along a single axis: \emph{when} natural language enters the agent's lifecycle and, consequently, \emph{what} it modifies. This yields three pillars, summarized in Figure~\ref{fig:taxonomy} and synthesized in Table~\ref{tab:pillar-overview}.

\begin{itemize}
    \setlength\itemsep{0.25em}
    \item \textbf{Pillar~1: Language as Grounding Signal} (\S\ref{sec:pillar1}) acts at \emph{problem-definition time}, specifying the MDP \citep{sutton1998reinforcement} itself: goals, states, actions, and rewards. A natural-language task description such as \textit{``You are in a room with two doors''} defines what the agent perceives, how it can act, and what counts as success, all before any policy runs.
    \item \textbf{Pillar~2: Language as Deliberative Feedback} (\S\ref{sec:pillar2}) acts at \emph{inference time}, refining outputs or reasoning traces through critique, memory, debate, or search, without updating parameters. A message such as \textit{``Step~3 has an error; try dividing by~2, not~4''} redirects a single generation episode but leaves the model unchanged for future tasks.
    \item \textbf{Pillar~3: Language as Learning Signal} (\S\ref{sec:pillar3}) acts at \emph{training time}, where verbal feedback is distilled into gradient updates that persistently reshape the policy. A judgment such as \textit{``Response~A is better than B because it addresses the edge case''} becomes a preference pair that alters model weights, affecting all subsequent interactions.
\end{itemize}

\noindent The key distinction is \emph{persistence}: grounding defines the task space, deliberative feedback improves a single episode, and learning signal permanently reshapes the policy. We discuss each pillar in consecutive sections. Methods spanning multiple pillars are categorized by their role in the primary function of verbal feedback.

\begin{table}[ht]
    \centering
    \small
    \renewcommand{\arraystretch}{1.2}
    \caption{Overview of the three VRL pillars.}
    \label{tab:pillar-overview}
    \begin{tabularx}{\linewidth}{@{} l *{3}{>{\raggedright\arraybackslash}X} @{}}
        \toprule
        \textbf{Axis} & \textbf{Grounding Signal} & \textbf{Deliberative Feedback} & \textbf{Learning Signal} \\
        \midrule
        \textbf{When} & Problem-definition time & Inference time & Training time \\
        \midrule
        \textbf{Modifies} & Task, State, Action, Reward & Outputs or reasoning traces & Model weights, Training data \\
        \midrule
        \textbf{Persistence} & Defines the task & Single episode (extendable via memory) & All future interactions \\
        \midrule
        \textbf{Methods} & Environment specification & Prompting and critique & SFT, PPO, DPO \\
        \midrule
        \textbf{Challenges} & Grounding gap; Compositionality & Feedback Quality; Compute Cost & Signal Compression; Filter Quality \\
        \bottomrule
    \end{tabularx}
\end{table}

\paragraph{Working Example: the coding agent - }
The three pillars are complementary, not competing; they operate together across a single development cycle. Consider an agent \citep{wang2025openhands} tasked with resolving a GitHub issue. \emph{Grounding} (Pillar~1) enters first: the natural-language issue description defines the goal, the repository structure specifies the state space, and the test suite implicitly encodes a reward function. At inference time, \emph{deliberative feedback} (Pillar~2) drives the iterative loop: the agent generates a patch, executes the tests, treats traces and error messages as verbal feedback, critiques its own output, and revises. When successful and failed trajectories are later collected and used to fine-tune the model via preference optimization, they become a \emph{learning signal} (Pillar~3) that persistently improves the policy for future tasks. The same agent thus drives reinforcement loop by consuming verbal feedback at three distinct timescales, each producing a different effect. This example demonstrates the rationale for organizing by temporal scope rather than by feedback source (human, model, tool) or modality (text, scalar, code). Identical verbal feedback can yield fundamentally different effects depending on when it is consumed, and each temporal regime introduces distinct technical challenges, which we will discuss in later sections.

\section{Pillar 1: Language as Grounding Signal}%
\label{sec:pillar1}

\begin{figure}[!t]
     \centering    
    \includegraphics[width=1\linewidth]{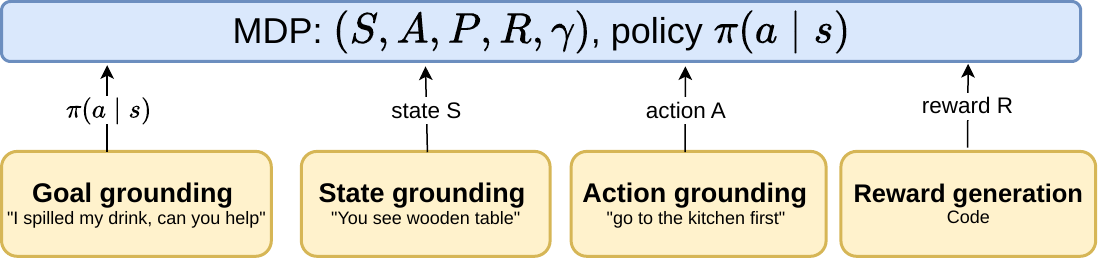}
    
    \caption{\footnotesize Grounding of MDP components through natural language. Each subcategory maps verbal feedback to a specific element of the MDP: goal grounding specifies the objective, state grounding represents observations as text, action grounding resolves instructions to executable behavior, and reward code generation compiles language into executable reward functions.}
    \label{fig:groundingsignal}
\end{figure}

In this pillar, we discuss the role of language in providing grounding for each component of the task in which an agent operates.As illustrated in Figure~\ref{fig:groundingsignal}, verbal feedback maps to specific MDP elements \citep{sutton1998reinforcement}: specifying what the agent should optimize (goals - \S\ref{sec:goal-grounding}), what it perceives (states- \S\ref{sec:state-grounding}), how it can act (actions- \S\ref{sec:action-grounding}), and what constitutes success (rewards- \S\ref{sec:reward-code}) \citep{ macglashan2017interactive}. This matters because many failures in language-based agents stem not from poor optimization but from poor grounding: feedback that cannot map to executable actions or valid reward conditions will not improve the agent. A common bottleneck across all four subcategories is the grounding gap. More detailed verbal specifications do not necessarily result in improved grounding. The critical factor is whether the mapping from language to MDP components is sufficiently precise for the agent to execute actions.
 
\subsection{Goal Grounding}%
\label{sec:goal-grounding}
Goal grounding uses verbal feedback to specify what the agent should achieve, requiring parsing instructions into objects, relations, and target conditions \citep{chaplot2018gated}, filtering plans by physical executability \citep{ahn2022can, driess2023palm}, and decomposing goals into verifiable subgoals. Recent work extends this to multi-agent coordination \citep{zhang2025towards}, hierarchical decomposition via offline RL \citep{hu2025divide}, and training from language specifications without manual reward functions \citep{li2026ground}. The open challenge is compositional generalization: agents must follow goals assembled from familiar primitives in unfamiliar combinations. BabyAI \citep{chevalier2018babyai} isolates this in gridworlds, and even state-of-the-art LLMs achieve only about 75\% coverage under compositional constraints \citep{sakai2025revisiting}, though compositional policy representations can improve sample efficiency on novel compositions \citep{cohen2025compositional}.

\subsection{State Grounding}%
\label{sec:state-grounding}
Where goal grounding defines what to achieve, state grounding defines what the agent perceives. Verbal state descriptions are most explicit in text-based environments \citep{yao2020keep, cote2018textworld}, but appear whenever visual or physical states are summarized into language. \citet{shridhar2020alfworld} show that agents trained on verbal state descriptions transfer to embodied 3D settings, demonstrating language as a modality-invariant state representation. Recent work includes closed-loop state feedback for robotic planning \citep{bhat2024grounding} and scene-graph-based grounding \citep{huang2025escacontextualizingembodiedagents}. A key challenge is information preservation, specifically whether the verbal abstraction retains sufficient signal for the policy to operate effectively.

\subsection{Action Grounding}%
\label{sec:action-grounding}
Given a goal and a perceived state, the agent must determine how to act. Action grounding resolves verbal instructions to specific skills, tool calls, or motor commands, requiring that instructions map to executable actions and that execution outcomes feedback as language to revise subsequent steps. \citet{huang2022inner} demonstrate this closed loop with Inner Monologue, while \citet{sharma2022correcting} show that short corrective utterances can amend mistaken steps without disrupting the rest of the plan. Recent systems have pushed toward tighter integration: multi-modal grounded replanning that corrects ambiguous instructions using visual cues \citep{kim2025multi}, constraint-aware visual programming for proactive failure detection \citep{zhou2025code}, corrective planning that handles physical, logical, and semantic errors across hierarchical control levels \citep{joublin2024copal}, and visually grounded task-and-motion planning that combines LLM reasoning with physical feasibility \citep{zhang2025llm}. The central challenge is granularity alignment: the same instruction can refer to a temporally extended skill (``make coffee''), a discrete subgoal (``go to the kitchen''), or a motor command (``close the gripper''), each demanding a different language-to-control interface.

\subsection{Reward Code Generation}%
\label{sec:reward-code}
The preceding subcategories ground objectives, observations, and actions in language. Reward code generation closes the loop by grounding success itself. This subcategory overlaps with Pillar~3 when generated rewards are used for training; we categorize it here because verbal feedback defines the reward function itself, becoming part of the MDP. The distinguishing feature is compiling verbal descriptions into executable reward functions rather than using language as a direct learning signal. \citet{ma2024eureka} introduced Eureka, where an LLM generates and iteratively refines reward code based on training summaries, and \citet{xie2024text2reward} extended this to dense reward shaping. This area has seen rapid recent progress: CARD \citep{sun2025large} introduces dynamic trajectory-based feedback to refine reward code without per-iteration RL training, PROF \citep{sun2025prof} extends the paradigm to offline settings through preference optimization, and \citet{zeng2024video2reward} show that reward functions can be generated from video demonstrations. To ensure effectiveness, verbal feedback must be compilable, executable, and accurately reflect the original natural-language intent.

\section{Pillar 2: Language as Deliberative Feedback}%
\label{sec:pillar2}

\begin{figure}[!t]
     \centering    
    \includegraphics[width=1.05\linewidth]{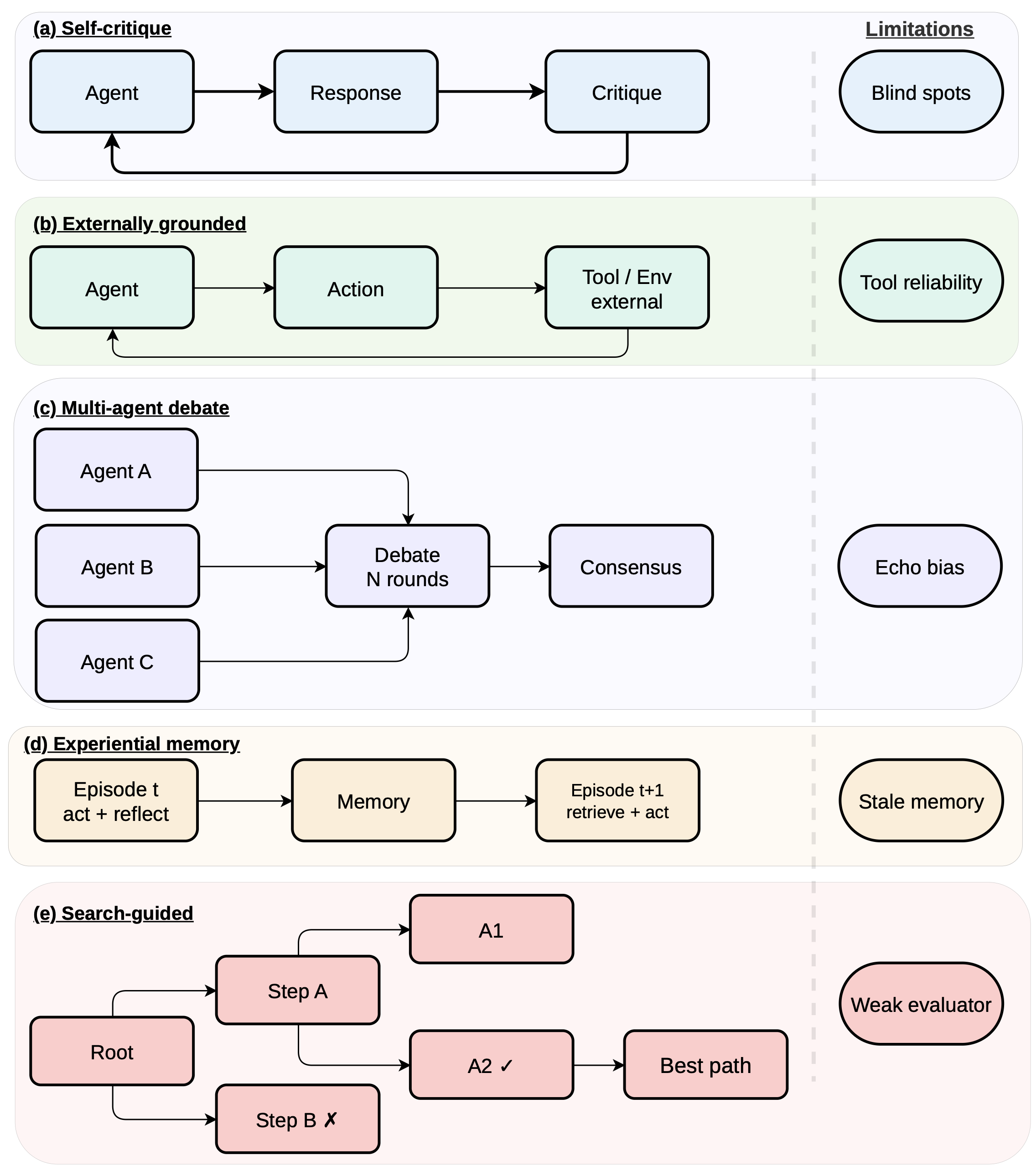}
    \caption{\footnotesize Inference-time verbal feedback mechanisms. Each subcategory exhibits a distinct feedback loop structure: (a) self-critique loops within a single model, (b) externally grounded critique incorporates verified tool or environment output, (c) multi-agent debate distributes critique across parallel model instances and converges to a refined output, (d) experiential memory persists lessons across episodes, and (e) search-guided deliberation explores and prunes multiple candidate paths.}
    \label{fig:inference_time}
\end{figure}

With the task grounded, we turn to methods that refine reasoning within a single episode without updating model parameters, closely aligning with test-time compute scaling \citep{snell2024scaling}. As illustrated in Figure~\ref{fig:inference_time}, five subcategories differ in where critique originates: the model itself (\S\ref{sec:self-refine}), external tools (\S\ref{sec:external-critique}), peer models (\S\ref{sec:multi-agent-critique}), past episodes (\S\ref{sec:experiential Memory}), or parallel search paths (\S\ref{sec:search-guided}).

\subsection{Self-Critique}%
\label{sec:self-refine}
The simplest deliberative loop asks the model to critique its own output and revise accordingly. Self-Refine \citep{madaan2023selfrefine} formalized this pattern, building on Constitutional AI's \citep{bai2022constitutional} demonstration that models can evaluate outputs against written principles. The primary challenge is circularity. Replacing self-generated critique with evaluation from a separately trained critic enhances reasoning \citep{paul2024refiner}, which indicates that feedback quality is the main bottleneck. Self-correction degrades when the critique source shares the generator's blind spots \citep{huang2023large, kamoi2024can}, especially in smaller models \citep{olausson2024self}. Mitigation strategies include Socratic decomposition into verifiable steps \citep{shi2025ssr} and parallel self-refinement across candidates \citep{wang2025learning}. However, when errors arise from gaps in the model's knowledge, internal feedback mechanisms cannot identify the missing information.

\subsection{Externally Grounded Critique}%
\label{sec:external-critique}

External grounding addresses this circularity by anchoring verbal feedback in deterministic tool output \citep{schick2023toolformer}. Verified signals, such as execution traces, unit test verdicts, search results, and API responses \citep{chen2024teaching, gou2024toratoolintegratedreasoningagent}, facilitate the generation of corrected outputs \citep{gou2024critic, yang2024swe, wang2024executable}. Additionally, end-to-end reinforcement learning enables models to utilize such feedback more effectively \citep{gehring2024rlef}. The primary failure mode shifts from blind spots to trust asymmetry: the model receives accurate tool output but may misattribute the root cause of a reported trace. Practical limitations persist, including infrastructure overhead, poor scalability of latency, and the restriction of grounded critique to domains with a verifiable oracle.

\subsection{Multi-Agent Critique / Debate}%
\label{sec:multi-agent-critique}
Multi-agent debate distributes verbal critique across parallel model instances, frequently assigning distinct personas to enhance perspective \citep{hong2024metagpt}. The exchange of critiques can improve reasoning even among identical models \citep{du2023debate}, while the introduction of diverse roles further strengthens the quality of discussion \citep{chan2023chatevalbetterllmbasedevaluators}. Communication topology is a critical factor; full-broadcast designs incur substantial token overhead \citep{choi2026debate}, motivating the exploration of sparse alternatives \citep{li2024improving}. Models may also be trained specifically for debate, enabling them to revise reasoning in response to conflicting opinions \citep{liu2026learningselfdebatepreparingreasoning}. Adaptive heterogeneous frameworks have demonstrated improvements in mathematical reasoning \citep{zhou2025adaptive}. However, the primary limitation is that distributing critique is only effective when participating models are genuinely diverse. When models share biases, the process yields redundant consensus \citep{estornell2024multi, liang2024encouragingdivergentthinkinglarge}, and current debate methods do not consistently outperform simpler single-agent strategies \citep{choi2026debate}.

\subsection{Experiential Memory}%
\label{sec:experiential Memory}

Previous approaches regenerate feedback from scratch in each episode. In contrast, experiential memory distills reusable lessons into persistent storage, categorized as episodic (raw logs), semantic (distilled rules), and procedural (reusable skills) \citep{zhang2025survey}. Reflexion \citep{shinn2023reflexion} stores raw verbal feedback, ExpeL \citep{zhao2024expel} extracts insights from paired successes and failures, and Voyager \citep{wang2023voyager} accumulates executable skills. Recent surveys have formalized these mechanisms \citep{du2026memory}, while MemBench evaluates memory quality through downstream performance \citep{tan2025membench}. However, persisting feedback can also perpetuate errors: poor-quality memories propagate mistakes, expanding memory stores require complex retrieval \citep{park2023genagents}, outdated knowledge can mislead in non-stationary environments \citep{zhang2025survey}, and adversarial injection may induce cross-session drift \citep{dong2026memory}.

\subsection{Search-Guided Deliberation}%
\label{sec:search-guided}
While previous approaches refine a single trajectory, search-guided deliberation explores multiple candidate paths and employs verbal feedback to determine which paths to expand, prune, or abandon \citep{wang-etal-2025-dont}. The Tree of Thoughts framework \citep{yao2023tree} samples several intermediate thoughts and evaluates them through self-assessment, thereby extending chain-of-thought prompting \citep{wei2022chain} with a tree-based search mechanism. Reasoning via Planning \citep{hao2023reasoning} and Graph of Thoughts \citep{besta2024graph} further generalize the underlying topology, permitting non-linear dependencies such as aggregation and refinement. A key advantage of these methods is that verbal feedback can be applied at every thought level, which facilitates earlier identification of unproductive paths. However, exploring multiple states necessitates numerous large language model (LLM) calls, rendering these methods considerably more expensive than single-pass generation \citep{xie2023self}.

\section{Pillar 3: Language as Learning Signal}%
\begin{figure}[t]
     \centering    
    \includegraphics[width=1.05\linewidth]{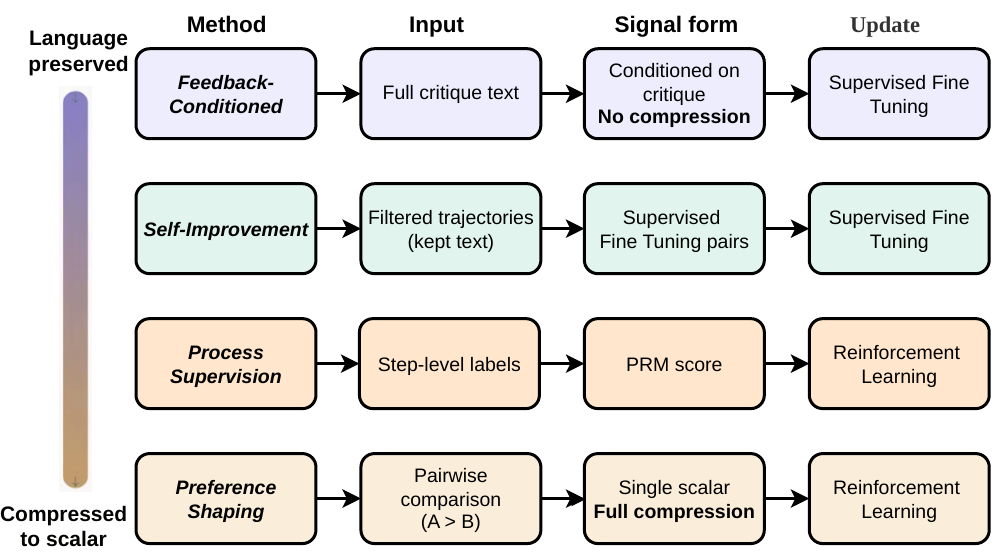}
    \caption{\footnotesize Training-time verbal feedback methods organized by degree of linguistic compression. Methods at the top preserve full natural-language feedback during training, while methods at the bottom compress it into scalar signals.}
    \label{fig:training_time}
\end{figure}

\label{sec:pillar3}
The method above leave the model unchanged; we now examine how verbal feedback can persistently reshape the policy through training. As illustrated in Figure~\ref{fig:training_time}, the four subcategories form a compression spectrum: from feedback-conditioned modeling (\S\ref{sec:feedback-conditioned}), which preserves full critiques, through self-improvement (\S\ref{sec:self-improvement}) and process supervision (\S\ref{sec:process-supervision}), to preference shaping (\S\ref{sec:preference-shaping}), which reduces verbal judgement to a single scalar.

\subsection{Feedback-Conditioned Modeling}%
\label{sec:feedback-conditioned}

At the upper end of the compression spectrum, feedback-conditioned modeling maintains verbal critiques in their entirety by conditioning the model directly on natural-language feedback during training. The training signal consists of a triple $(x, v, a^*)$: the input, a critique of an initial response, and the revised output. A central question is whether to retain the critique within the training context. Early studies omitted the critique, using only the refined output as the supervision target \citep{scheurer2023training}. Subsequent work, such as ALT \citep{lloret2024towards}, demonstrated that retaining feedback as conditioning context enables models to learn explicit mappings from critique to correction. \citet{luo2025fcp} further formalized this approach by treating free-form feedback as a primary conditioning signal with online bootstrapping. While preserving the full linguistic content of the critique allows these methods to retain the richest signal, this same receptiveness introduces a risk: models may learn to follow critiques indiscriminately, even when they are incorrect \citep{sharma2024towards}.

\subsection{Self-Improvement}
\label{sec:self-improvement}
Self-improvement approaches treat language as filtered text rather than as a conditioning context. The central mechanism is generate-then-filter: the model generates multiple candidate trajectories, verbal feedback signals determine which candidates are retained, and these selected outputs become supervised fine-tuning data for subsequent iterations \citep{singh2023beyond}. The filtering criterion directly shapes the model's learning. For example, in STaR \citep{zelikman2022star}, correctness of the final answer is used as the filter, while in Constitutional AI \citep{bai2022constitutional}, adherence to safety principles serves this role. Explicit labeling is not always required; majority voting across diverse samples can identify likely correct traces \citep{huang2022selfimprove}, and the model itself can act as a judge across iterations \citep{yuan2024self}. This self-contained loop can facilitate compounding improvements over multiple cycles. However, filter quality remains a critical bottleneck: when the same model functions as both generator and judge, systematic biases may be amplified \citep{shumailov2023curse}.

\subsection{Process Supervision}%
\label{sec:process-supervision}
Process supervision reduces verbal feedback to scalar scores at the step level. Annotators assess each intermediate reasoning step, and these evaluations are used to train a process reward model; the verbal content of the judgments is omitted. Despite this reduction, step-level supervision substantially outperforms outcome-level supervision \citep{lightman2024let, uesato2022solving}, as localized feedback enables more precise credit assignment. The high cost of step-level annotation has motivated efforts to automate this process \citep{wang2024mathshepherdverifyreinforcellms, luo2024improve}. Generative process reward models that incorporate reasoning prior to scoring retain more verbal information while maintaining step-level granularity \citep{zhao2025genprm, liu2025agenticreinforcementlearningimplicit}. Although step-level granularity offers significant advantages, it remains costly and is challenging to define in domains where intermediate correctness is ambiguous.

\subsection{Preference Shaping}%
\label{sec:preference-shaping}

Preference shaping represents the maximal compression of verbal feedback, reducing a comparative judgment between two responses to a single scalar that guides policy updates. Building on the foundation of learning from verbal preferences \citep{christiano2017deep}, InstructGPT \citep{ouyang2022training} demonstrated large-scale implementation by training a reward model on pairwise rankings and optimizing the policy online. Direct Preference Optimization (DPO) \citep{rafailov2023direct} further simplified this process by eliminating the explicit reward model and training directly on preference pairs, using the comparison itself as the loss signal. Subsequent research has investigated additional alternatives \citep{ethayarajh2024kto, hong2024orpo, meng2024simpo, azar2024general}. Notably, \citet{lee2023rlaif} replaced human annotators with AI judges, thereby enhancing the scalability of preference collection. Although scalar compression reduces the explanatory richness of verbal feedback, the scalability enabled by llm generated judgments increasingly justifies this tradeoff.

\section{Insights, Discussion and Future Prospects}
\label{sec:discussion}
In practice, the three pillars operate together, as the coding-agent example in \S\ref{sec:taxonomy} illustrates. Four cross-cutting challenges span them: feedback-model quality (\S\ref{sec:Feedback_models}), tool-interface design (\S\ref{sec:designing_tools}), adversarial robustness (\S\ref{sec:adversarial_verbal_feedback}), and the absence of formal guarantees (\S\ref{sec:theoritical_foundation}).

\subsection{Dedicated Feedback Models}
\label{sec:Feedback_models}
Verbal feedback quality remains the key performance factor, but current practice typically reuses the same LLM for both generation and critique. Dedicated critics consistently outperform this approach: Shepherd \citep{wang2023shepherd} provides more actionable feedback than untuned baselines, CriticGPT \citep{mcaleese2024criticgpt} detects code bugs more reliably, and CritiqueLLM \citep{ke2024critiquellm} scales critique data via multi-path prompting. Co-training critics with reasoners yields further improvements \citep{liu2026generativeadversarialreasonerenhancing}, while generative process reward models that reason before scoring preserve more verbal signal than scalar PRMs \citep{khalifa2025processrewardmodelsthink}. CriticBench \citep{lin2024criticbench} offers the first systematic benchmark for critique quality. These results indicate a natural next step: just as the field invested in instruction-tuned models \citep{ouyang2022training}, it should now develop feedback-tuned models whose objectives address error localization, actionability, and calibration. Recent advances in span-level feedback gradients \citep{wang2026text2gradreinforcementlearningnatural} and multi-turn didactic interactions \citep{klissarov2026improving} suggest such models can serve as differentiable training signals across each pillars.

\subsection{Designing Tools for Verbal Feedback}
\label{sec:designing_tools}
Developer-oriented tools often yield outputs that are terse or conflate error types. \citet{yang2024swe} demonstrated that LLM-optimized interfaces enhance agent performance, while \citet{bandlamudi2025framework} showed ambiguous outputs cause many agent failures across 750 API calls. Tool-interactive critiquing requires interpretable output \citep{gou2024critic}, and executable code actions benefit agents by providing structured feedback \citep{wang2024executable}. RL on execution feedback \citep{gehring2024rlef} presumes adequate tool-output signal, while agent harness frameworks \citep{ning2026code} set tool quality as the limit of iterative refinement. However, a metric to measure this limit is lacking: tool-output compliance should quantify agent consumability. Tool APIs must expose structured metadata distinguishing task-level errors from infrastructure failures to enable agents to apply suitable recovery strategies.

\subsection{Adversarial Verbal Feedback}
\label{sec:adversarial_verbal_feedback}
Because VRL agents inherently act on feedback, adversarial feedback constitutes policy manipulation, introducing vulnerabilities beyond standard prompt injection. \citet{zhan2025adaptive} demonstrated that adaptive attacks circumvent all eight tested defenses, while tool-level injection attains a 96.7\% success rate on GPT-4o \citep{shi2025prompt}. Hidden instructions in API responses covertly manipulate agents \citep{zhan2024injecagent}, adversarial memory injection induces cross-session drift \citep{dong2026memory}, and even benign incorrect feedback leads to sycophancy \citep{sharma2024towards}. Defenses are developing: instruction hierarchies \citep{wallace2024instruction}, adversarial fine-tuning \citep{chen2025secaligndefendingpromptinjection}, and system-level design patterns \citep{beurer2025design} each mitigate narrow attack surfaces. Comprehensive solutions will require feedback-provenance mechanisms to verify source integrity and assign trust, as well as adversarial feedback benchmarks to evaluate robustness as the primary metric.

\subsection{Theoretical Foundations}
\label{sec:theoritical_foundation}
Verbal RL lacks formal guarantees; it remains unclear when verbal feedback enhances sample efficiency or policy quality. Three promising directions emerge. First, if verbal critics serve as noisy oracles with bounded error, PAC-learning frameworks \citep{strehl2009reinforcement} can bound VRL sample complexity, and critic benchmarks \citep{lin2024criticbench} allow empirical estimation of error rates. Second, modeling feedback as partial observation links VRL to POMDP instruction following \citep{luketina2019survey}, where belief-state planning yields convergence guarantees; \citet{mccallum2023feedback} implement this with feedback-conditioned decision transformers. Third, rate-distortion theory may quantify signal retention from critique to scalar, clarifying when preference shaping aligns with feedback-conditioned methods. Empirical studies support these directions: language models learn from verbal feedback without scalar rewards \citep{luo2025fcp}, text feedback broadens RL capabilities \citep{song2026expanding}, and recent surveys link RLHF to language-conditioned RL theory \citep{kaufmann2023survey}. Formalizing results in these areas would clarify when the cost of rich verbal feedback is warranted over scalar alternatives.

\section{Conclusion}
\label{sec:conclusion}
In this paper, we present the first comprehensive survey of verbal reinforcement learning, organized around three pillars based on when language enters an agent's lifecycle: grounding signal, deliberative feedback, and learning signal. Together, these pillars reveal a common pattern: verbal feedback is rapidly becoming the primary medium for defining, updating, and improving agents.

Looking ahead, we anticipate that verbal feedback will become a first-class
design primitive in agent architectures. The boundaries between pillars will
increasingly blur as agents consume verbal feedback in unified loops that span
grounding, deliberation, and learning within a single trajectory. The bottleneck
will shift from \textit{generating} feedback to \textit{verifying} it, making
feedback provenance, quality benchmarks, and adversarial robustness first-class
infrastructure requirements. Finally, tool interfaces will need to be redesigned
around agent consumability rather than human readability, as the gap between
tool output quality and agent performance becomes the practical ceiling for
iterative refinement.

\newpage
\section*{Limitations}
We aim to provide broad coverage of methods that use verbal feedback within the agent's lifecycle. Within each category, we focus on representative papers rather than attempting an exhaustive enumeration. Additionally, while our taxonomy offers a unifying framework, it may oversimplify methods that span multiple roles of language. Several approaches do not fit neatly into a single branch, since verbal feedback can simultaneously serve as a grounding signal, a deliberative mechanism, and a learning signal. Our taxonomy should therefore be understood as an organizing lens rather than a strict partition of the literature. Finally, our survey focuses primarily on methods in which verbal feedback is explicit and central to the agent’s process, and therefore does not fully cover adjacent work where language plays a more auxiliary role. AI writing tools were used for language editing and proofreading only; all research contributions are the authors' own.

\bibliography{references}

@inproceedings{wang2024executable,
  title={Executable code actions elicit better llm agents},
  author={Wang, Xingyao and Chen, Yangyi and Yuan, Lifan and Zhang, Yizhe and Li, Yunzhu and Peng, Hao and Ji, Heng},
  booktitle={Forty-first International Conference on Machine Learning},
  year={2024}
}

@article{strehl2009reinforcement,
  title={Reinforcement learning in finite MDPs: PAC analysis.},
  author={Strehl, Alexander L and Li, Lihong and Littman, Michael L},
  journal={Journal of Machine Learning Research},
  volume={10},
  number={11},
  year={2009}
}

@article{wallace2024instruction,
  title={The instruction hierarchy: Training llms to prioritize privileged instructions},
  author={Wallace, Eric and Xiao, Kai and Leike, Reimar and Weng, Lilian and Heidecke, Johannes and Beutel, Alex},
  journal={arXiv preprint arXiv:2404.13208},
  year={2024}
}

@inproceedings{zhan2024injecagent,
  title={Injecagent: Benchmarking indirect prompt injections in tool-integrated large language model agents},
  author={Zhan, Qiusi and Liang, Zhixiang and Ying, Zifan and Kang, Daniel},
  booktitle={Findings of the Association for Computational Linguistics: ACL 2024},
  pages={10471--10506},
  year={2024}
}

@article{lu2024ai,
  title={The ai scientist: Towards fully automated open-ended scientific discovery},
  author={Lu, Chris and Lu, Cong and Lange, Robert Tjarko and Foerster, Jakob and Clune, Jeff and Ha, David},
  journal={arXiv preprint arXiv:2408.06292},
  year={2024}
}

@article{m2024augmenting,
  title={Augmenting large language models with chemistry tools},
  author={Bran, Andres and Cox, Sam and Schilter, Oliver and Baldassari, Carlo and White, Andrew D and Schwaller, Philippe},
  journal={Nature machine intelligence},
  volume={6},
  number={5},
  pages={525--535},
  year={2024},
  publisher={Nature Publishing Group UK London}
}

@inproceedings{olausson2024self,
  title={Is self-repair a silver bullet for code generation?},
  author={Olausson, Theo X and Inala, Jeevana Priya and Wang, Chenglong and Gao, Jianfeng and Solar-Lezama, Armando},
  booktitle={International Conference on Learning Representations},
  volume={2024},
  pages={36545--36593},
  year={2024}
}

@misc{jimenez2024swebenchlanguagemodelsresolve,
      title={SWE-bench: Can Language Models Resolve Real-World GitHub Issues?}, 
      author={Carlos E. Jimenez and John Yang and Alexander Wettig and Shunyu Yao and Kexin Pei and Ofir Press and Karthik Narasimhan},
      year={2024},
      eprint={2310.06770},
      archivePrefix={arXiv},
      primaryClass={cs.CL},
      url={https://arxiv.org/abs/2310.06770}, 
}

@inproceedings{sharma2024towards,
  title={Towards understanding sycophancy in language models},
  author={Sharma, Mrinank and Tong, Meg and Korbak, Tomek and Duvenaud, David and Askell, Amanda and Bowman, Sam and Durmus, Esin and Hatfield-Dodds, Zac and Johnston, Scott and Kravec, Shauna},
  booktitle={International Conference on Learning Representations},
  volume={2024},
  pages={110--144},
  year={2024}
}

@inproceedings{lloret2024towards,
  title={Towards aligning language models with textual feedback},
  author={Lloret, Sa{\"u}c Abadal and Dhuliawala, Shehzaad and Murugesan, Keerthiram and Sachan, Mrinmaya},
  booktitle={Proceedings of the 2024 Conference on Empirical Methods in Natural Language Processing},
  pages={20240--20266},
  year={2024}
}

@article{luo2024improve,
  title={Improve mathematical reasoning in language models by automated process supervision},
  author={Luo, Liangchen and Liu, Yinxiao and Liu, Rosanne and Phatale, Samrat and Guo, Meiqi and Lara, Harsh and Li, Yunxuan and Shu, Lei and Zhu, Yun and Meng, Lei},
  journal={arXiv preprint arXiv:2406.06592},
  year={2024}
}

@misc{wang2024mathshepherdverifyreinforcellms,
      title={Math-Shepherd: Verify and Reinforce LLMs Step-by-step without Human Annotations}, 
      author={Peiyi Wang and Lei Li and Zhihong Shao and R. X. Xu and Damai Dai and Yifei Li and Deli Chen and Y. Wu and Zhifang Sui},
      year={2024},
      eprint={2312.08935},
      archivePrefix={arXiv},
      primaryClass={cs.AI},
      url={https://arxiv.org/abs/2312.08935}, 
}

@article{uesato2022solving,
  title={Solving math word problems with process-and outcome-based feedback},
  author={Uesato, Jonathan and Kushman, Nate and Kumar, Ramana and Song, Francis and Siegel, Noah and Wang, Lisa and Creswell, Antonia and Irving, Geoffrey and Higgins, Irina},
  journal={arXiv preprint arXiv:2211.14275},
  year={2022}
}

@article{shumailov2023curse,
  title={The curse of recursion: Training on generated data makes models forget},
  author={Shumailov, Ilia and Shumaylov, Zakhar and Zhao, Yiren and Gal, Yarin and Papernot, Nicolas and Anderson, Ross},
  journal={arXiv preprint arXiv:2305.17493},
  year={2023}
}

@inproceedings{huang2023large,
  title={Large language models cannot self-correct reasoning yet},
  author={Huang, Jie and Chen, Xinyun and Mishra, Swaroop and Zheng, Huaixiu Steven and Yu, Adams and Song, Xinying and Zhou, Denny},
  booktitle={International conference on learning representations},
  volume={2024},
  pages={32808--32824},
  year={2024}
}

@article{achiam2023gpt,
  title={Gpt-4 technical report},
  author={Achiam, Josh and Adler, Steven and Agarwal, Sandhini and Ahmad, Lama and Akkaya, Ilge and Aleman, Florencia Leoni and Almeida, Diogo and Altenschmidt, Janko and Altman, Sam and Anadkat, Shyamal},
  journal={arXiv preprint arXiv:2303.08774},
  year={2023}
}

@inproceedings{wang-etal-2025-dont,
    title = "Don{'}t Get Lost in the Trees: Streamlining {LLM} Reasoning by Overcoming Tree Search Exploration Pitfalls",
    author = "Wang, Ante  and
      Song, Linfeng  and
      Tian, Ye  and
      Yu, Dian  and
      Mi, Haitao  and
      Duan, Xiangyu  and
      Tu, Zhaopeng  and
      Su, Jinsong  and
      Yu, Dong",
    editor = "Che, Wanxiang  and
      Nabende, Joyce  and
      Shutova, Ekaterina  and
      Pilehvar, Mohammad Taher",
    booktitle = "Proceedings of the 63rd Annual Meeting of the Association for Computational Linguistics (Volume 1: Long Papers)",
    month = jul,
    year = "2025",
    address = "Vienna, Austria",
    publisher = "Association for Computational Linguistics",
    url = "https://aclanthology.org/2025.acl-long.1167/",
    doi = "10.18653/v1/2025.acl-long.1167",
    pages = "23946--23959",
    ISBN = "979-8-89176-251-0"
}

@article{xie2023self,
  title={Self-evaluation guided beam search for reasoning},
  author={Xie, Yuxi and Kawaguchi, Kenji and Zhao, Yiran and Zhao, James Xu and Kan, Min-Yen and He, Junxian and Xie, Michael},
  journal={Advances in Neural Information Processing Systems},
  volume={36},
  pages={41618--41650},
  year={2023}
}

@article{zhang2025survey,
  title={A survey on the memory mechanism of large language model-based agents},
  author={Zhang, Zeyu and Dai, Quanyu and Bo, Xiaohe and Ma, Chen and Li, Rui and Chen, Xu and Zhu, Jieming and Dong, Zhenhua and Wen, Ji-Rong},
  journal={ACM Transactions on Information Systems},
  volume={43},
  number={6},
  pages={1--47},
  year={2025},
  publisher={ACM New York, NY}
}

@inproceedings{li2024improving,
  title={Improving multi-agent debate with sparse communication topology},
  author={Li, Yunxuan and Du, Yibing and Zhang, Jiageng and Hou, Le and Grabowski, Peter and Li, Yeqing and Ie, Eugene},
  booktitle={Findings of the Association for Computational Linguistics: EMNLP 2024},
  pages={7281--7294},
  year={2024}
}

@article{choi2026debate,
  title={Debate or vote: Which yields better decisions in multi-agent large language models?},
  author={Choi, Hyeong Kyu and Zhu, Jerry and Li, Sharon},
  journal={Advances in Neural Information Processing Systems},
  volume={38},
  pages={101732--101764},
  year={2026}
}

@article{estornell2024multi,
  title={Multi-llm debate: Framework, principals, and interventions},
  author={Estornell, Andrew and Liu, Yang},
  journal={Advances in Neural Information Processing Systems},
  volume={37},
  pages={28938--28964},
  year={2024}
}

@misc{chan2023chatevalbetterllmbasedevaluators,
      title={ChatEval: Towards Better LLM-based Evaluators through Multi-Agent Debate}, 
      author={Chi-Min Chan and Weize Chen and Yusheng Su and Jianxuan Yu and Wei Xue and Shanghang Zhang and Jie Fu and Zhiyuan Liu},
      year={2023},
      eprint={2308.07201},
      archivePrefix={arXiv},
      primaryClass={cs.CL},
      url={https://arxiv.org/abs/2308.07201}, 
}

@misc{liang2024encouragingdivergentthinkinglarge,
      title={Encouraging Divergent Thinking in Large Language Models through Multi-Agent Debate}, 
      author={Tian Liang and Zhiwei He and Wenxiang Jiao and Xing Wang and Yan Wang and Rui Wang and Yujiu Yang and Shuming Shi and Zhaopeng Tu},
      year={2024},
      eprint={2305.19118},
      archivePrefix={arXiv},
      primaryClass={cs.CL},
      url={https://arxiv.org/abs/2305.19118}, 
}

@article{yang2024swe,
  title={Swe-agent: Agent-computer interfaces enable automated software engineering},
  author={Yang, John and Jimenez, Carlos E and Wettig, Alexander and Lieret, Kilian and Yao, Shunyu and Narasimhan, Karthik and Press, Ofir},
  journal={Advances in Neural Information Processing Systems},
  volume={37},
  pages={50528--50652},
  year={2024}
}

@misc{gou2024toratoolintegratedreasoningagent,
      title={ToRA: A Tool-Integrated Reasoning Agent for Mathematical Problem Solving}, 
      author={Zhibin Gou and Zhihong Shao and Yeyun Gong and Yelong Shen and Yujiu Yang and Minlie Huang and Nan Duan and Weizhu Chen},
      year={2024},
      eprint={2309.17452},
      archivePrefix={arXiv},
      primaryClass={cs.CL},
      url={https://arxiv.org/abs/2309.17452}, 
}

@inproceedings{chen2020simple,
  title={A simple framework for contrastive learning of visual representations},
  author={Chen, Ting and Kornblith, Simon and Norouzi, Mohammad and Hinton, Geoffrey},
  booktitle={International conference on machine learning},
  pages={1597--1607},
  year={2020},
  organization={PmLR}
}

@misc{devlin2019bertpretrainingdeepbidirectional,
      title={BERT: Pre-training of Deep Bidirectional Transformers for Language Understanding}, 
      author={Jacob Devlin and Ming-Wei Chang and Kenton Lee and Kristina Toutanova},
      year={2019},
      eprint={1810.04805},
      archivePrefix={arXiv},
      primaryClass={cs.CL},
      url={https://arxiv.org/abs/1810.04805}, 
}

@article{brown2020language,
  title={Language models are few-shot learners},
  author={Brown, Tom and Mann, Benjamin and Ryder, Nick and Subbiah, Melanie and Kaplan, Jared D and Dhariwal, Prafulla and Neelakantan, Arvind and Shyam, Pranav and Sastry, Girish and Askell, Amanda},
  journal={Advances in neural information processing systems},
  volume={33},
  pages={1877--1901},
  year={2020}
}

@article{mnih2015human,
  title={Human-level control through deep reinforcement learning},
  author={Mnih, Volodymyr and Kavukcuoglu, Koray and Silver, David and Rusu, Andrei A and Veness, Joel and Bellemare, Marc G and Graves, Alex and Riedmiller, Martin and Fidjeland, Andreas K and Ostrovski, Georg},
  journal={Nature},
  volume={518},
  number={7540},
  pages={529--533},
  year={2015}
}

@misc{vamplew2021scalarrewardenoughresponse,
      title={Scalar reward is not enough: A response to Silver, Singh, Precup and Sutton (2021)}, 
      author={Peter Vamplew and Benjamin J. Smith and Johan Kallstrom and Gabriel Ramos and Roxana Radulescu and Diederik M. Roijers and Conor F. Hayes and Fredrik Heintz and Patrick Mannion and Pieter J. K. Libin and Richard Dazeley and Cameron Foale},
      year={2021},
      eprint={2112.15422},
      archivePrefix={arXiv},
      primaryClass={cs.AI},
      url={https://arxiv.org/abs/2112.15422}, 
}

@article{amodei2016concrete,
  title={Concrete problems in AI safety},
  author={Amodei, Dario and Olah, Chris and Steinhardt, Jacob and Christiano, Paul and Schulman, John and Man{\'e}, Dan},
  journal={arXiv preprint arXiv:1606.06565},
  year={2016}
}

@article{degrave2022magnetic,
  title={Magnetic control of tokamak plasmas through deep reinforcement learning},
  author={Degrave, Jonas and Felici, Federico and Buchli, Jonas and Neunert, Michael and Tracey, Brendan and Carpanese, Francesco and Ewalds, Timo and Hafner, Roland and Abdolmaleki, Abbas and de Las Casas, Diego},
  journal={Nature},
  volume={602},
  number={7897},
  pages={414--419},
  year={2022},
  publisher={Nature Publishing Group UK London}
}

@article{mirhoseini2021graph,
  title={A graph placement methodology for fast chip design},
  author={Mirhoseini, Azalia and Goldie, Anna and Yazgan, Mustafa and Jiang, Joe Wenjie and Songhori, Ebrahim and Wang, Shen and Lee, Young-Joon and Johnson, Eric and Pathak, Omkar and Nova, Azade},
  journal={Nature},
  volume={594},
  number={7862},
  pages={207--212},
  year={2021},
  publisher={Nature Publishing Group UK London}
}

@article{silver2016mastering,
  title={Mastering the game of Go with deep neural networks and tree search},
  author={Silver, David and Huang, Aja and Maddison, Chris J and Guez, Arthur and Sifre, Laurent and Van Den Driessche, George and Schrittwieser, Julian and Antonoglou, Ioannis and Panneershelvam, Veda and Lanctot, Marc},
  journal={nature},
  volume={529},
  number={7587},
  pages={484--489},
  year={2016},
  publisher={Nature Publishing Group UK London}
}

@article{yao2023tree,
  title={Tree of thoughts: Deliberate problem solving with large language models},
  author={Yao, Shunyu and Yu, Dian and Zhao, Jeffrey and Shafran, Izhak and Griffiths, Tom and Cao, Yuan and Narasimhan, Karthik},
  journal={Advances in neural information processing systems},
  volume={36},
  pages={11809--11822},
  year={2023}
}

@inproceedings{shinn2023reflexion,
  title={Reflexion: Language Agents with Verbal Reinforcement Learning},
  author={Shinn, Noah and Cassano, Federico and Labash, Berman and Gopinath, Ashwin and Narasimhan, Karthik and Yao, Shunyu},
  booktitle={Advances in Neural Information Processing Systems},
  volume={36},
  year={2023}
}

@inproceedings{madaan2023selfrefine,
  title={Self-Refine: Iterative Refinement with Self-Feedback},
  author={Madaan, Aman and Tandon, Niket and Gupta, Prakhar and Hallinan, Skyler and Gao, Luyu and Wiegreffe, Sarah and Alon, Uri and Dziri, Nouha and Prabhumoye, Shrimai and Yang, Yiming and Gupta, Shashank and Majumder, Bodhisattwa Prasad and Hermann, Katherine and Welleck, Sean and Yazdanbakhsh, Amir and Clark, Peter},
  booktitle={Advances in Neural Information Processing Systems},
  year={2023}
}

@article{ouyang2022training,
  title={Training language models to follow instructions with human feedback},
  author={Ouyang, Long and Wu, Jeffrey and Jiang, Xu and Almeida, Diogo and Wainwright, Carroll and Mishkin, Pamela and Zhang, Chong and Agarwal, Sandhini and Slama, Katarina and Ray, Alex},
  journal={Advances in neural information processing systems},
  volume={35},
  pages={27730--27744},
  year={2022}
}

@article{christiano2017deep,
  title={Deep reinforcement learning from human preferences},
  author={Christiano, Paul F and Leike, Jan and Brown, Tom and Martic, Miljan and Legg, Shane and Amodei, Dario},
  journal={Advances in Neural Information Processing Systems},
  volume={30},
  year={2017}
}

@inproceedings{gou2024critic,
  title={{CRITIC}: Large Language Models Can Self-Correct with Tool-Interactive Critiquing},
  author={Gou, Zhibin and Shao, Zhihong and Gong, Yeyun and Shen, Yelong and Yang, Yujiu and Duan, Nan and Chen, Weizhu},
  booktitle={International Conference on Learning Representations},
  year={2024}
}

@inproceedings{cote2018textworld,
  title={Textworld: A learning environment for text-based games},
  author={C{\^o}t{\'e}, Marc-Alexandre and K{\'a}d{\'a}r, Akos and Yuan, Xingdi and Kybartas, Ben and Barnes, Tavian and Fine, Emery and Moore, James and Hausknecht, Matthew and El Asri, Layla and Adada, Mahmoud},
  booktitle={Workshop on Computer Games},
  pages={41--75},
  year={2018},
  organization={Springer}
}

@article{driess2023palm,
  title={Palm-e: An embodied multimodal language model},
  author={Driess, Danny and Xia, Fei and Sajjadi, Mehdi SM and Lynch, Corey and Chowdhery, Aakanksha and Ichter, Brian and Wahid, Ayzaan and Tompson, Jonathan and Vuong, Quan and Yu, Tianhe},
  journal={arXiv preprint arXiv:2303.03378},
  year={2023}
}

@inproceedings{chaplot2018gated,
  title={Gated-attention architectures for task-oriented language grounding},
  author={Chaplot, Devendra Singh and Sathyendra, Kanthashree Mysore and Pasumarthi, Rama Kumar and Rajagopal, Dheeraj and Salakhutdinov, Ruslan},
  booktitle={Proceedings of the AAAI Conference on Artificial Intelligence},
  volume={32},
  year={2018}
}

@inproceedings{macglashan2017interactive,
  title={Interactive learning from policy-dependent human feedback},
  author={MacGlashan, James and Ho, Mark K and Loftin, Robert and Peng, Bei and Wang, Guan and Roberts, David L and Taylor, Matthew E and Littman, Michael L},
  booktitle={International conference on machine learning},
  pages={2285--2294},
  year={2017},
  organization={PMLR}
}

@article{lee2023rlaif,
  title={Rlaif vs. rlhf: Scaling reinforcement learning from human feedback with ai feedback},
  author={Lee, Harrison and Phatale, Samrat and Mansoor, Hassan and Mesnard, Thomas and Ferret, Johan and Lu, Kellie and Bishop, Colton and Hall, Ethan and Carbune, Victor and Rastogi, Abhinav},
  journal={arXiv preprint arXiv:2309.00267},
  year={2023}
}

@article{singh2023beyond,
  title={Beyond human data: Scaling self-training for problem-solving with language models},
  author={Singh, Avi and Co-Reyes, John D and Agarwal, Rishabh and Anand, Ankesh and Patil, Piyush and Garcia, Xavier and Liu, Peter J and Harrison, James and Lee, Jaehoon and Xu, Kelvin},
  journal={arXiv preprint arXiv:2312.06585},
  year={2023}
}

@article{meng2024simpo,
  title={Simpo: Simple preference optimization with a reference-free reward},
  author={Meng, Yu and Xia, Mengzhou and Chen, Danqi},
  journal={Advances in Neural Information Processing Systems},
  volume={37},
  pages={124198--124235},
  year={2024}
}

@inproceedings{hong2024orpo,
  title={Orpo: Monolithic preference optimization without reference model},
  author={Hong, Jiwoo and Lee, Noah and Thorne, James},
  booktitle={Proceedings of the 2024 Conference on Empirical Methods in Natural Language Processing},
  pages={11170--11189},
  year={2024}
}

@inproceedings{azar2024general,
  title={A general theoretical paradigm to understand learning from human preferences},
  author={Azar, Mohammad Gheshlaghi and Guo, Zhaohan Daniel and Piot, Bilal and Munos, Remi and Rowland, Mark and Valko, Michal and Calandriello, Daniele},
  booktitle={International Conference on Artificial Intelligence and Statistics},
  pages={4447--4455},
  year={2024},
  organization={PMLR}
}

@article{ethayarajh2024kto,
  title={Kto: Model alignment as prospect theoretic optimization},
  author={Ethayarajh, Kawin and Xu, Winnie and Muennighoff, Niklas and Jurafsky, Dan and Kiela, Douwe},
  journal={arXiv preprint arXiv:2402.01306},
  year={2024}
}

@article{yuan2024self,
  title={Self-rewarding language models},
  author={Yuan, Weizhe and Pang, Richard Yuanzhe and Cho, Kyunghyun and Li, Xian and Sukhbaatar, Sainbayar and Xu, Jing and Weston, Jason},
  journal={arXiv preprint arXiv:2401.10020},
  year={2024}
}

@inproceedings{lightman2024let,
  title={Let's verify step by step},
  author={Lightman, Hunter and Kosaraju, Vineet and Burda, Yuri and Edwards, Harrison and Baker, Bowen and Lee, Teddy and Leike, Jan and Schulman, John and Sutskever, Ilya and Cobbe, Karl},
  booktitle={International Conference on Learning Representations},
  volume={2024},
  pages={39578--39601},
  year={2024}
}

@inproceedings{hong2024metagpt,
  title={MetaGPT: Meta programming for a multi-agent collaborative framework},
  author={Hong, Sirui and Zhuge, Mingchen and Chen, Jonathan and Zheng, Xiawu and Cheng, Yuheng and Wang, Jinlin and Zhang, Ceyao and Yau, Steven and Lin, Zijuan and Zhou, Liyang },
  booktitle={International Conference on Learning Representations},
  volume={2024},
  pages={23247--23275},
  year={2024}
}

@inproceedings{besta2024graph,
  title={Graph of thoughts: Solving elaborate problems with large language models},
  author={Besta, Maciej and Blach, Nils and Kubicek, Ales and Gerstenberger, Robert and Podstawski, Michal and Gianinazzi, Lukas and Gajda, Joanna and Lehmann, Tomasz and Niewiadomski, Hubert and Nyczyk, Piotr},
  booktitle={Proceedings of the AAAI conference on artificial intelligence},
  volume={38},
  pages={17682--17690},
  year={2024}
}

@article{schick2023toolformer,
  title={Toolformer: Language models can teach themselves to use tools},
  author={Schick, Timo and Dwivedi-Yu, Jane and Dess{\`\i}, Roberto and Raileanu, Roberta and Lomeli, Maria and Hambro, Eric and Zettlemoyer, Luke and Cancedda, Nicola and Scialom, Thomas},
  journal={Advances in neural information processing systems},
  volume={36},
  pages={68539--68551},
  year={2023}
}

@inproceedings{chen2024teaching,
  title={Teaching large language models to self-debug},
  author={Chen, Xinyun and Lin, Maxwell and Sch{\"a}rli, Nathanael and Zhou, Denny},
  booktitle={International Conference on Learning Representations},
  volume={2024},
  pages={8746--8825},
  year={2024}
}

@article{wei2022chain,
  title={Chain-of-thought prompting elicits reasoning in large language models},
  author={Wei, Jason and Wang, Xuezhi and Schuurmans, Dale and Bosma, Maarten and Xia, Fei and Chi, Ed and Le, Quoc V and Zhou, Denny },
  journal={Advances in neural information processing systems},
  volume={35},
  pages={24824--24837},
  year={2022}
}

@article{kamoi2024can,
  title={When can llms actually correct their own mistakes? a critical survey of self-correction of llms},
  author={Kamoi, Ryo and Zhang, Yusen and Zhang, Nan and Han, Jiawei and Zhang, Rui},
  journal={Transactions of the Association for Computational Linguistics},
  volume={12},
  pages={1417--1440},
  year={2024}
}

@article{pternea2024rl,
  title={The rl/llm taxonomy tree: Reviewing synergies between reinforcement learning and large language models},
  author={Pternea, Moschoula and Singh, Prerna and Chakraborty, Abir and Oruganti, Yagna and Milletari, Mirco and Bapat, Sayli and Jiang, Kebei},
  journal={Journal of Artificial Intelligence Research},
  volume={80},
  pages={1525--1573},
  year={2024}
}

@article{luketina2019survey,
  title={A survey of reinforcement learning informed by natural language},
  author={Luketina, Jelena and Nardelli, Nantas and Farquhar, Gregory and Foerster, Jakob and Andreas, Jacob and Grefenstette, Edward and Whiteson, Shimon and Rockt{\"a}schel, Tim},
  journal={arXiv preprint arXiv:1906.03926},
  year={2019}
}

@inproceedings{xie2024text2reward,
  title={Text2reward: Reward shaping with language models for reinforcement learning},
  author={Xie, Tianbao and Zhao, Siheng and Wu, Chen and Liu, Yitao and Luo, Qian and Zhong, Victor and Yang, Yanchao and Yu, Tao},
  booktitle={International Conference on Learning Representations},
  volume={2024},
  pages={35663--35699},
  year={2024}
}

@inproceedings{ma2024eureka,
  title={Eureka: Human-level reward design via coding large language models},
  author={Ma, Yecheng Jason and Liang, William and Wang, Guanzhi and Huang, De-An and Bastani, Osbert and Jayaraman, Dinesh and Zhu, Yuke and Fan, Jim },
  booktitle={International conference on learning Representations},
  volume={2024},
  pages={26516--26560},
  year={2024}
}

@article{sharma2022correcting,
  title={Correcting robot plans with natural language feedback},
  author={Sharma, Pratyusha and Sundaralingam, Balakumar and Blukis, Valts and Paxton, Chris and Hermans, Tucker and Torralba, Antonio and Andreas, Jacob and Fox, Dieter},
  journal={arXiv preprint arXiv:2204.05186},
  year={2022}
}

@article{huang2022inner,
  title={Inner monologue: Embodied reasoning through planning with language models},
  author={Huang, Wenlong and Xia, Fei and Xiao, Ted and Chan, Harris and Liang, Jacky and Florence, Pete and Zeng, Andy and Tompson, Jonathan and Mordatch, Igor and Chebotar, Yevgen},
  journal={arXiv preprint arXiv:2207.05608},
  year={2022}
}

@article{shridhar2020alfworld,
  title={Alfworld: Aligning text and embodied environments for interactive learning},
  author={Shridhar, Mohit and Yuan, Xingdi and C{\^o}t{\'e}, Marc-Alexandre and Bisk, Yonatan and Trischler, Adam and Hausknecht, Matthew},
  journal={arXiv preprint arXiv:2010.03768},
  year={2020}
}

@inproceedings{yao2020keep,
  title={Keep calm and explore: Language models for action generation in text-based games},
  author={Yao, Shunyu and Rao, Rohan and Hausknecht, Matthew and Narasimhan, Karthik},
  booktitle={Proceedings of the 2020 Conference on Empirical Methods in Natural Language Processing (EMNLP)},
  pages={8736--8754},
  year={2020}
}

@article{chevalier2018babyai,
  title={Babyai: A platform to study the sample efficiency of grounded language learning},
  author={Chevalier-Boisvert, Maxime and Bahdanau, Dzmitry and Lahlou, Salem and Willems, Lucas and Saharia, Chitwan and Nguyen, Thien Huu and Bengio, Yoshua},
  journal={arXiv preprint arXiv:1810.08272},
  year={2018}
}

@article{ahn2022can,
  title={Do as i can, not as i say: Grounding language in robotic affordances},
  author={Ahn, Michael and Brohan, Anthony and Brown, Noah and Chebotar, Yevgen and Cortes, Omar and David, Byron and Finn, Chelsea and Fu, Chuyuan and Gopalakrishnan, Keerthana and Hausman, Karol},
  journal={arXiv preprint arXiv:2204.01691},
  year={2022}
}

@inproceedings{du2023debate,
  title={Improving Factuality and Reasoning in Language Models through Multiagent Debate},
  author={Du, Yilun and Li, Shuang and Torralba, Antonio and Tenenbaum, Joshua B and Mordatch, Igor},
  booktitle={International Conference on Machine Learning},
  year={2024}
}

@article{bai2022constitutional,
  title={Constitutional {AI}: Harmlessness from {AI} Feedback},
  author={Bai, Yuntao and Kadavath, Saurav and Kundu, Sandipan and Askell, Amanda and Kernion, Jackson and Jones, Andy and Chen, Anna and Goldie, Anna and Mirhoseini, Azalia and McKinnon, Cameron and Chen, Carol and Olsson, Catherine and Olah, Christopher and Hernandez, Danny and Drain, Dawn and Ganguli, Deep and Li, Dustin and Tran-Johnson, Eli and Perez, Ethan and Kerr, Jamie and Mueller, Jared and Ladish, Jeffrey and Landau, Joshua and Ndousse, Kamal and Lukosuite, Kamile and Lovitt, Liane and Sellitto, Michael and Elhage, Nelson and Schiefer, Nicholas and Mercado, Noemi and DasSarma, Nova and Lasenby, Robert and Larson, Robin and Ringer, Sam and Johnston, Scott and Kravec, Shauna and Showk, Sheer El and Fort, Stanislav and Lanham, Tamera and Telleen-Lawton, Timothy and Conerly, Tom and Henighan, Tom and Hume, Tristan and Bowman, Samuel R and Hatfield-Dodds, Zac and Mann, Ben and Amodei, Dario and Joseph, Nicholas and McCandlish, Sam and Brown, Tom and Kaplan, Jared},
  journal={arXiv preprint arXiv:2212.08073},
  year={2022}
}

@article{luo2025fcp,
  title={Language models can learn from verbal feedback without scalar rewards},
  author={Luo, Renjie and Liu, Zichen and Liu, Xiangyan and Du, Chao and Lin, Min and Chen, Wenhu and Lu, Wei and Pang, Tianyu},
  journal={arXiv preprint arXiv:2509.22638},
  year={2025}
}

@inproceedings{hao2023reasoning,
  title={Reasoning with language model is planning with world model},
  author={Hao, Shibo and Gu, Yi and Ma, Haodi and Hong, Joshua and Wang, Zhen and Wang, Daisy and Hu, Zhiting},
  booktitle={Proceedings of the 2023 Conference on Empirical Methods in Natural Language Processing},
  pages={8154--8173},
  year={2023}
}

@article{paul2024refiner,
  title={{REFINER}: Reasoning Feedback on Intermediate Representations},
  author={Paul, Debjit and Ismayilzada, Mete and Peyrard, Maxime and Borges, Beatriz and Bosselut, Antoine and West, Robert and Faltings, Boi},
  journal={arXiv preprint arXiv:2304.01904},
  year={2024}
}

@article{scheurer2023training,
  title={Training language models with language feedback at scale},
  author={Scheurer, J{\'e}r{\'e}my and Campos, Jon Ander and Korbak, Tomasz and Chan, Jun Shern and Chen, Angelica and Cho, Kyunghyun and Perez, Ethan},
  journal={arXiv preprint arXiv:2303.16755},
  year={2023}
}

@article{huang2022selfimprove,
  title={Large Language Models Can Self-Improve},
  author={Huang, Jiaxin and Gu, Shixiang Shane and Hou, Le and Wu, Yuexin and Wang, Xuezhi and Yu, Hongkun and Han, Jiawei},
  journal={arXiv preprint arXiv:2210.11610},
  year={2022}
}

@article{rafailov2023direct,
  title={Direct preference optimization: Your language model is secretly a reward model},
  author={Rafailov, Rafael and Sharma, Archit and Mitchell, Eric and Manning, Christopher D and Ermon, Stefano and Finn, Chelsea},
  journal={Advances in neural information processing systems},
  volume={36},
  pages={53728--53741},
  year={2023}
}

@inproceedings{zelikman2022star,
  title={{STaR}: Bootstrapping Reasoning With Reasoning},
  author={Zelikman, Eric and Wu, Yuhuai and Mu, Jesse and Goodman, Noah D},
  booktitle={Advances in Neural Information Processing Systems},
  year={2022}
}

@inproceedings{zhao2025genprm,
  title={Genprm: Scaling test-time compute of process reward models via generative reasoning},
  author={Zhao, Jian and Liu, Runze and Zhang, Kaiyan and Zhou, Zhimu and Gao, Junqi and Li, Dong and Lyu, Jiafei and Qian, Zhouyi and Qi, Biqing and Li, Xiu},
  booktitle={Proceedings of the AAAI Conference on Artificial Intelligence},
  volume={40},
  number={41},
  pages={34932--34940},
  year={2026}
}

@article{wang2023shepherd,
  title={Shepherd: A Critic for Language Model Generation},
  author={Wang, Tianlu and Yu, Ping and Tan, Xiaoqing Ellen and O'Brien, Sean and Pasunuru, Ramakanth and Dwivedi-Yu, Jane and Golovneva, Olga and Zettlemoyer, Luke and Fazel-Zarandi, Maryam and Celikyilmaz, Asli},
  journal={arXiv preprint arXiv:2308.04592},
  year={2023}
}

@article{mcaleese2024criticgpt,
  title={{LLM} Critics Help Catch {LLM} Bugs},
  author={McAleese, Nat and Pokorny, Rai Michael and Uribe, Juan Felipe Ceron and Nitishinskaya, Evgenia and Trebacz, Maja and Leike, Jan},
  journal={arXiv preprint arXiv:2407.00215},
  year={2024}
}

@article{wang2023voyager,
  title={Voyager: An Open-Ended Embodied Agent with Large Language Models},
  author={Wang, Guanzhi and Xie, Yuqi and Jiang, Yunfan and Mandlekar, Ajay and Xiao, Chaowei and Zhu, Yuke and Fan, Linxi and Anandkumar, Anima},
  journal={arXiv preprint arXiv:2305.16291},
  year={2023}
}

@inproceedings{park2023genagents,
  title={Generative Agents: Interactive Simulacra of Human Behavior},
  author={Park, Joon Sung and O'Brien, Joseph and Cai, Carrie Jun and Morris, Meredith Ringel and Liang, Percy and Bernstein, Michael S},
  booktitle={ACM Symposium on User Interface Software and Technology (UIST)},
  year={2023}
}

@inproceedings{zhao2024expel,
  title={{ExpeL}: {LLM} Agents Are Experiential Learners},
  author={Zhao, Andrew and Huang, Daniel and Xu, Quentin and Lin, Matthieu and Liu, Yong-Jin and Huang, Gao},
  booktitle={AAAI Conference on Artificial Intelligence},
  year={2024}
}

@article{snell2024scaling,
  title={Scaling {LLM} Test-Time Compute Optimally Can Be More Effective than Scaling Model Parameters},
  author={Snell, Charlie and Lee, Jaehoon and Xu, Kelvin and Kumar, Aviral},
  journal={arXiv preprint arXiv:2408.03314},
  year={2024}
}

@article{ning2026code,
  title={Code as Agent Harness},
  author={Ning, Xuying and Tieu, Katherine and Fu, Dongqi and Wei, Tianxin and Li, Zihao and Bei, Yuanchen and Zou, Jiaru and Ai, Mengting and Liu, Zhining and Li, Ting-Wei},
  journal={arXiv preprint arXiv:2605.18747},
  year={2026}
}

@article{cohen2025compositional,
  title={Compositional Instruction Following with Language Models and Reinforcement Learning},
  author={Cohen, Vanya and Tasse, Geraud Nangue and Gopalan, Nakul and James, Steven and Gombolay, Matthew and Mooney, Ray and Rosman, Benjamin},
  journal={arXiv preprint arXiv:2501.12539},
  year={2025}
}

@article{li2026ground,
  title={Ground-Compose-Reinforce: Grounding Language in Agentic Behaviours using Limited Data},
  author={Li, Andrew and Klassen, Toryn and Wang, Andrew and A Alamdari, Parand and McIlraith, Sheila},
  journal={Advances in Neural Information Processing Systems},
  volume={38},
  pages={160838--160874},
  year={2026}
}

@article{hu2025divide,
  title={Divide and conquer: Grounding LLMs as efficient decision-making agents via offline hierarchical reinforcement learning},
  author={Hu, Zican and Liu, Wei and Qu, Xiaoye and Yue, Xiangyu and Chen, Chunlin and Wang, Zhi and Cheng, Yu},
  journal={arXiv preprint arXiv:2505.19761},
  year={2025}
}

@inproceedings{zhang2025towards,
  title={Towards efficient llm grounding for embodied multi-agent collaboration},
  author={Zhang, Yang and Yang, Shixin and Bai, Chenjia and Wu, Fei and Li, Xiu and Wang, Zhen and Li, Xuelong},
  booktitle={Findings of the Association for Computational Linguistics: ACL 2025},
  pages={1663--1699},
  year={2025}
}

@inproceedings{sakai2025revisiting,
  title={Revisiting compositional generalization capability of large language models considering instruction following ability},
  author={Sakai, Yusuke and Kamigaito, Hidetaka and Watanabe, Taro},
  booktitle={Proceedings of the 63rd Annual Meeting of the Association for Computational Linguistics (Volume 1: Long Papers)},
  pages={31219--31238},
  year={2025}
}

@article{bhat2024grounding,
  title={Grounding llms for robot task planning using closed-loop state feedback},
  author={Bhat, Vineet and Kaypak, Ali Umut and Krishnamurthy, Prashanth and Karri, Ramesh and Khorrami, Farshad},
  journal={arXiv preprint arXiv:2402.08546},
  year={2024}
}

@misc{huang2025escacontextualizingembodiedagents,
      title={ESCA: Contextualizing Embodied Agents via Scene-Graph Generation}, 
      author={Jiani Huang and Amish Sethi and Matthew Kuo and Mayank Keoliya and Neelay Velingker and JungHo Jung and Ser-Nam Lim and Ziyang Li and Mayur Naik},
      year={2025},
      eprint={2510.15963},
      archivePrefix={arXiv},
      primaryClass={cs.CV},
      url={https://arxiv.org/abs/2510.15963}, 
}

@inproceedings{zhou2025code,
  title={Code-as-monitor: Constraint-aware visual programming for reactive and proactive robotic failure detection},
  author={Zhou, Enshen and Su, Qi and Chi, Cheng and Zhang, Zhizheng and Wang, Zhongyuan and Huang, Tiejun and Sheng, Lu and Wang, He},
  booktitle={Proceedings of the Computer Vision and Pattern Recognition Conference},
  pages={6919--6929},
  year={2025}
}

@inproceedings{kim2025multi,
  title={Multi-modal grounded planning and efficient replanning for learning embodied agents with a few examples},
  author={Kim, Taewoong and Kim, Byeonghwi and Choi, Jonghyun},
  booktitle={Proceedings of the AAAI Conference on Artificial Intelligence},
  volume={39},
  number={4},
  pages={4329--4337},
  year={2025}
}

@article{zhang2025llm,
  title={LLM-GROP: Visually grounded robot task and motion planning with large language models},
  author={Zhang, Xiaohan and Ding, Yan and Hayamizu, Yohei and Altaweel, Zainab and Zhu, Yifeng and Zhu, Yuke and Stone, Peter and Paxton, Chris and Zhang, Shiqi},
  journal={The International Journal of Robotics Research},
  pages={02783649251378196},
  year={2025},
  publisher={SAGE Publications Sage UK: London, England}
}

@inproceedings{joublin2024copal,
  title={Copal: corrective planning of robot actions with large language models},
  author={Joublin, Frank and Ceravola, Antonello and Smirnov, Pavel and Ocker, Felix and Deigmoeller, Joerg and Belardinelli, Anna and Wang, Chao and Hasler, Stephan and Tanneberg, Daniel and Gienger, Michael},
  booktitle={2024 ieee international conference on robotics and automation (ICRA)},
  pages={8664--8670},
  year={2024},
  organization={IEEE}
}

@article{sun2025large,
  title={A large language model-driven reward design framework via dynamic feedback for reinforcement learning},
  author={Sun, Shengjie and Liu, Runze and Lyu, Jiafei and Yang, Jing-Wen and Zhang, Liangpeng and Li, Xiu},
  journal={Knowledge-Based Systems},
  volume={326},
  pages={114065},
  year={2025},
  publisher={Elsevier}
}

@article{sun2025prof,
  title={PROF: An LLM-based Reward Code Preference Optimization Framework for Offline Imitation Learning},
  author={Sun, Shengjie and Lyu, Jiafei and Liu, Runze and Yan, Mengbei and Liu, Bo and Ye, Deheng and Li, Xiu},
  journal={arXiv preprint arXiv:2511.13765},
  year={2025}
}

@article{zeng2024video2reward,
  title={Video2reward: Generating reward function from videos for legged robot behavior learning},
  author={Zeng, Runhao and Zhou, Dingjie and Liang, Qiwei and Liu, Junlin and Li, Hui and Huang, Changxin and Li, Jianqiang and Hu, Xiping and Sun, Fuchun},
  journal={arXiv preprint arXiv:2412.05515},
  year={2024}
}

@article{shi2025ssr,
  title={SSR: Socratic Self-Refine for Large Language Model Reasoning},
  author={Shi, Haizhou and Liu, Ye and Pang, Bo and Liu, Zeyu Leo and Wang, Hao and Savarese, Silvio and Xiong, Caiming and Zhou, Yingbo and Yavuz, Semih},
  journal={arXiv preprint arXiv:2511.10621},
  year={2025}
}

@article{wang2025learning,
  title={Learning to refine: Self-refinement of parallel reasoning in llms},
  author={Wang, Qibin and Zhao, Pu and Huang, Shaohan and Yang, Fangkai and Wang, Lu and Wei, Furu and Lin, Qingwei and Rajmohan, Saravan and Zhang, Dongmei},
  journal={arXiv preprint arXiv:2509.00084},
  year={2025}
}

@article{gehring2024rlef,
  title={Rlef: Grounding code llms in execution feedback with reinforcement learning},
  author={Gehring, Jonas and Zheng, Kunhao and Copet, Jade and Mella, Vegard and Carbonneaux, Quentin and Cohen, Taco and Synnaeve, Gabriel},
  journal={arXiv preprint arXiv:2410.02089},
  year={2024}
}

@misc{liu2026learningselfdebatepreparingreasoning,
      title={Learning from Self-Debate: Preparing Reasoning Models for Multi-Agent Debate}, 
      author={Chenxi Liu and Yanshuo Chen and Ruibo Chen and Tianyi Xiong and Tong Zheng and Heng Huang},
      year={2026},
      eprint={2601.22297},
      archivePrefix={arXiv},
      primaryClass={cs.CL},
      url={https://arxiv.org/abs/2601.22297}, 
}

@article{zhou2025adaptive,
  title={Adaptive heterogeneous multi-agent debate for enhanced educational and factual reasoning in large language models},
  author={Zhou, Yan and Chen, Yanguang},
  journal={Journal of King Saud University Computer and Information Sciences},
  volume={37},
  number={10},
  pages={330},
  year={2025},
  publisher={Springer}
}

@article{du2026memory,
  title={Memory for autonomous llm agents: Mechanisms, evaluation, and emerging frontiers},
  author={Du, Pengfei},
  journal={arXiv preprint arXiv:2603.07670},
  year={2026}
}

@inproceedings{tan2025membench,
  title={Membench: Towards more comprehensive evaluation on the memory of llm-based agents},
  author={Tan, Haoran and Zhang, Zeyu and Ma, Chen and Chen, Xu and Dai, Quanyu and Dong, Zhenhua},
  booktitle={Findings of the Association for Computational Linguistics: ACL 2025},
  pages={19336--19352},
  year={2025}
}

@article{dong2026memory,
  title={Memory injection attacks on LLM agents via query-only interaction},
  author={Dong, Shen and Xu, Shaochen and He, Pengfei and Li, Yige and Tang, Jiliang and Liu, Tianming and Liu, Hui and Xiang, Zhen},
  journal={Advances in Neural Information Processing Systems},
  volume={38},
  pages={46697--46731},
  year={2026}
}

@article{klissarov2026improving,
  title={Improving Interactive In-Context Learning from Natural Language Feedback},
  author={Klissarov, Martin and Cook, Jonathan and Antognini, Diego and Sun, Hao and Li, Jingling and Jaques, Natasha and Musat, Claudiu and Grefenstette, Edward},
  journal={arXiv preprint arXiv:2602.16066},
  year={2026}
}

@misc{liu2025agenticreinforcementlearningimplicit,
      title={Agentic Reinforcement Learning with Implicit Step Rewards}, 
      author={Xiaoqian Liu and Ke Wang and Yuchuan Wu and Fei Huang and Yongbin Li and Junge Zhang and Jianbin Jiao},
      year={2025},
      eprint={2509.19199},
      archivePrefix={arXiv},
      primaryClass={cs.CL},
      url={https://arxiv.org/abs/2509.19199}, 
}

\newpage
\appendix

\end{document}